\documentclass{llncs}
\usepackage{amsmath}
\usepackage{times}
\usepackage{amssymb}
\usepackage{enumerate}
\usepackage{amsfonts}
\usepackage{graphicx}
\usepackage{xspace}
\usepackage{xstring}
\usepackage{mathrsfs}
\usepackage{color}
\usepackage{url}
\usepackage[latin1]{inputenc}
\usepackage[hidelinks]{hyperref}

\newcounter{instr}

\title{Prior Polarity Lexical Resources\\for the Italian Language
\thanks{This work has been accepted for publication at the 12th International Workshop on Natural Language Processing and Cognitive Science (NLPCS 2015).}
}

\author{
Valeria Borz\`i$^{\dagger}$, Simone Faro$^{\dagger,\ast}$, Arianna Pavone$^{\ddagger}$ and Sabrina Sansone$^{\ddagger}$}
\institute{
	$^{\dagger}$Dipartimento di Matematica e Informatica, Universit\`a di Catania\\Viale Andrea Doria 6, I-95125 Catania, Italy\\[0.2cm]
	$^{\ddagger}$Dipartimento di Scienze Umanistiche, Universit\`a di Catania\\Piazza Dante 32, I-95124 Catania, Italy\\[0.2cm]
	\email{$^{\ast}$faro@dmi.unict.it}
 }

\newcommand{\rsname}{\textsc{Sabrina}\xspace}
\newcommand{\rsdescription}{Sentiment Analysis: a Broad Resource for Italian Natural language Applications\xspace}

\newcommand{\expl}[1]{{\color{blue}\StrSubstitute[0]{#1}{ }{+}[\word]\href{http://www.dmi.unict.it/~faro/sabrina/index.php?frase=\word}{\emph{#1}}}}

\begin{document}

\maketitle

\begin{abstract}
In this paper we present \rsname (\rsdescription) a manually annotated prior polarity lexical resource for Italian natural language applications in the field of opinion mining and sentiment induction. The resource consists in two different sets, an Italian dictionary of more than 277.000 words tagged with their prior polarity value, and a set of polarity modifiers, containing more than 200 words, which can be used in combination with non neutral terms of the dictionary in order to induce the sentiment of Italian compound terms.
To the best of our knowledge this is the first prior polarity manually annotated resource which has been developed for the Italian natural language.
\end{abstract}

\section{Introduction}
Sentiment classification~\cite{BL12,LZ12,MHK14,BFP14} concerns the use of automatic approaches for predicting the orientation of subjective content on text documents, with applications on many areas including information retrieval, customer intelligence and recommender and advertising systems.
Such discipline, where sentiment, opinion or emotion, are identified and classified in human written text is well known as \emph{sentiment analysis}. 

With the rapid increase of available subjective text on the internet in the form of blog posts, comments in discussion forums and product reviews, mining the user's opinion can assist in a lot of potential applications in areas such as recommender systems, search engines and market research.

Although some attempts have been made to extend solutions to other languages, till date all research efforts found in sentiment analysis literature deal mostly with English texts. However, in order to identify sentiment from a text, a lexical analysis of the source language plays a crucial role.

An approach for detecting sentiment in texts concerns the use of \emph{lexical resources} such as a dictionaries of opinionated terms. For example the terms \emph{love}, \emph{good} and \emph{favorite}  directly indicate a positive sentiment or an opinion, while words like \emph{hate}, \emph{bad} and \emph{scandal} can be associated with a negative sentiment.

Among the others, SentiWordNet~\cite{ES06} is one of the most used resource, containing opinion information on terms extracted from the WordNet database~\cite{Miller95} and made publicly available for research purposes. 
It is built via a semi supervised method and is considered a valuable resource for performing opinion mining tasks, providing a readily available database of term sentiment information for the English language. 

Other previous works~\cite{PLV02,ES06} have been already proposed for making dictionaries for those sentiment words using automatic approaches, however automatic identification of polarity orientation of such words is also a difficult research issue, known as \emph{polarity identification}.
In this context, it has been shown that the use of sentiment lexicons only provide a good baseline i.e. without using any natural language techniques only dictionary based approach produce a good performance~\cite{DB10}.

%

An alternative to automatic tagged resources are manually annotated lexicons which turns out to be undoubtedly more trustable although they took long time to be constructed and may be subject it annotator bias.

In this paper we present \rsname (\rsdescription) a manually annotated prior polarity lexical resource for Italian natural language applications in the field of opinion mining and sentiment induction. The resource consists in two different sets, an Italian dictionary of more than 277.000 words tagged with their prior polarity value, and a set of polarity modifiers, containing more than 200 words, which can be used in combination with non neutral terms of the dictionary in order to induce the sentiment of Italian compound terms.
To the best of our knowledge this is the first prior polarity manually annotated resource which has been developed for the Italian natural language.

The paper is organized as follows. In Section \ref{sec:polarity} we introduce the concept of prior and posterior polarity and present some known lexicons which label terms with their sentiment polarity. Then in Section \ref{sec:italian} we present the new tagged resources which has been created for the Italian language and discuss its properties. In Section \ref{sec:frontend} we briefly introduce also a web based fronted for accessing the resources. We draw our conclusions in Section \ref{sec:conclusions}.


\section{Prior and Posterior Polarity} \label{sec:polarity}
A typical computational approach to sentiment analysis starts with prior polarity lexicons where entries are tagged with their prior out of context polarity as human beings perceive using cognitive knowledge.

The \emph{prior polarity} of a term is the sentiment (positive or negative) that such word evokes by itself. 
More specifically we could define the prior polarity of a term as the polarity for its non-disambiguated meaning, out of any context.

For example the adjective \emph{cold} evokes (in most cases) a fairly negative sentiment, since it is used in sentences as \emph{a cold man}, \emph{a cold winter} or \emph{I feel cold}. However, depending on the context, we can find such term in sentences with a positive acceptation, as in \emph{I love cold beer}. 

In contrast with the prior polarity of a word, the polarities associated to each word sense is called in literature \emph{posterior polarity}.

In most cases prior polarity lexicons are lists of positive and negative words, often deployed as baselines or as features for other methods for sentiment analysis research~\cite{LZ12}. In these lexicon, words are associated with their prior polarity. 
For example it is presumable that the term \emph{wonderful} is associated with positive connotation while the term \emph{horrible} is associated with negative one. 
These approaches have the advantage of not needing deep semantic analysis or word sense disambiguation to assign an affective score to a word and are domain independent. In other word they are less precise but more portable. 


\subsection{Polarity Lexicons} \label{sec:lexicons}

Opinion lexicons are resources that associate sentiment orientation and words. Their use in opinion mining research stems from the hypothesis that individual words can be considered as a unit of opinion information, and therefore may provide clues to document sentiment and subjectivity.
These techniques could be broadly categorized in two genres: manual annotation  and automatic extraction of word polarity.

\begin{itemize}
\item \textbf{Manual annotation}.
Manual annotated lexicons are undoubtedly trustable but it took long time and, for these reasons, tend to be constrained to a small number of terms. By its nature, building manual lists is a time consuming effort, and may be subject to annotator bias.
Although such limitations manually created opinion lexicons were applied to sentiment classification as seen in~\cite{PLV02}, where a prediction of document polarity is given by counting positive and negative terms.\\


\item \textbf{Automatic detection}.
To overcome the above issues lexical induction approaches have been proposed in the literature with a view to extend the size of opinion lexicons from a core set of seed terms, either by exploring term relationships, or by evaluating similarities in document corpora. 
Early work in this area seen in~\cite{HM97} extends a list of positive and negative adjectives by evaluating conjunctive statements in a document corpus. 
However in most cases automatic processes still demands manual validations and, moreover, may fail to cover the multiple domains as automatic processes trust on specific corpus.
\end{itemize}

SentiWordNet~\cite{ES06} is one of the most popular lexical resources in Sentiment Analysis. It has been widely adopted since it provides a broad-coverage lexicon, built in a semi-automatic manner, for English  providing posterior polarities scores for each term of the language. 
It is the result of the automatic annotation of all the synsets of WordNet~\cite{Miller95} according to the notions of positivity, negativity, and neutrality. 
Different senses of the same term may thus have different opinion-related properties.



However in most  opinion mining applications it is necessary to derive prior polarities from the posteriors. 
For this reason several formulae to compute prior polarities starting from posterior polarities scores have been proposed in the literature. However, their performance varies significantly depending on the adopted variant.


For instance SentiWords is an inducted  prior polarity lexicon with the higher coverage for the English language. It contains roughly 155.000 words associated with a sentiment score included between $-1$ (strongly negative) and $1$ (strongly positive), learned from SentiWordNet. Words in this resource are also aligned with WordNet lists.  

For the sake of completeness we notice also that other prior polarity sentiment lexicons are available for the English language, such as Subjectivity Word List~\cite{Wilson05}, WordNet Affect list~\cite{Strapparava04} and the Taboada's adjective list~\cite{Voll07}.\\[0.2cm]

Although most of the efforts in literature have been devoted to the construction on lexicons resource for the English language, in recent years some research endeavors could be found in literature for solving the opinion mining problem in several languages and domains~\cite{DB10}. 
Until date most of the approaches to sentiment analysis in languages different from English consists in applying a word-translation from the target language to English before polarity extraction, which is applied by using one of the above described lexicons.
Such solutions, however, presents several problems including translation precision and disambiguation of words.

Recently some efforts have also been made to produce polarity lexicons also for languages different from English.
For instance in~\cite{DB10a} the authors proposed multiple computational techniques like, WordNet based, dictionary based, corpus based or generative approaches for generating SentiWordNet for Indian languages. 

For the sake of completeness we mention also an interactive gaming approach used for obtaining polarity values of english words, presented in~\cite{DB10}, where the authors proposed a tool, named Dr. Sentiment. to create and validate SentiWordNet in 56 languages by involving Internet population. 



\section{New broad Lexical Resources for the Italian Language} \label{sec:italian}
In this section we present \rsname\footnote{A tool for evaluating \rsname is available at the anonymous url \url{http://www.dmi.unict.it/~faro/sabrina}. Each example which you can find in this section is tagged with an anchor which redirect the reader to the web page of the tool in order to evaluate the sentiment value of the example itself.} (\rsdescription) a manually annotated prior polarity lexical resource for Italian natural language applications in the field of opinion mining and sentiment induction. The resource consists in two different sets, an Italian dictionary of more than 277.000 words tagged with their prior polarity value, and a set of polarity modifiers, containing more than 200 words, which can be used in combination with non neutral terms of the dictionary in order to induce the sentiment of Italian compound terms.

In recent years sentiment analysis in Italian texts has attracted attention due to Evalita, an initiative devoted to the evaluation of Natural Language Processing and Speech tools for Italian. In the recent Evalita 2014 edition the Sentipolc (SENTIment POLarity Classification) task\footnote{\url{http://www.di.unito.it/~tutreeb/sentipolc-evalita14/}} \cite{BBNPR14} was proposed.  It focused on Italian texts from Twitter by launching a battery of related tasks with an increasing level of complexity.

A first automatic annotated lexicon for the Italian language has been developed in \cite{BN13}, where the authors exploited three existing resources, namely MultiWordNet \cite{CMPS94}, SentiWordNet \cite{ES06}, and WordNet \cite{Miller95} to obtain an annotated lexicon of senses for Italian. It was named Sentix and basically port the SentiWordNet annotation to the Italian portion of MultiWordNet in a completely automatic fashion.
Sentix was then used in \cite{CCDB14} where the author described the UNITOR system that participated to the Sentipolc task within the context of Evalita 2014. The system has been developed as a workflow of Support Vector Machine classifiers. Specific features and kernel functions have been used to tackle the different sub-tasks, i.e. Subjectivity Classification, Polarity Classification and the pilot task Irony Detection.

To the best of our knowledge, besides Sentix, \rsname is the first prior polarity manually annotated resource which has been developed for the Italian natural language.


\subsection{Italian Polarity Lexicon}

Most sentiment lexicons in literature contain lists of tagged lemmas, i.e.  the canonical form ( or dictionary form) of a word. 
For instance the lastest version of MultiWordNet (1.39) contains around $58,000$ Italian word senses and $41,500$ lemmas organized into $32,700$ synsets aligned whenever possible with Princeton WordNet English synsets.
In using such kind of resources in sentiment analysis it is necessary to operate a previous step of sense disambiguation in order to identify the correspondent lemma of a word.

Our lexicon contains $277.387$ words of the Italian language, including their inflection, used in order to express different grammatical categories such as tense, mood, person, gender, etc.
For instance the dictionary contains the verb \emph{correre} (\emph{to run}) and its conjugations \emph{correvo}, \emph{correrà}, \emph{corressi}, etc.

Such set of words have been manually tagged with their prior polarities.
The annotation process started from the word set in the Ispell Italian dictionary\footnote{Ispell is a program that helps you to correct spelling and typographical errors in a file. When presented with a word that is not in the dictionary, ispell attempts to find near misses that might include the word you meant.} used for spell-checking purpose.
Each word of the lexicon has been associated with a polarity in the range between $-1$ and $1$, where $-1$ indicates a strongly negative polarity while $1$ indicates a very positive polarity. Mildly negative or positive opinion polarity have been tagged, respectively, with values $-0.5$ and $0.5$. In addition terms with a neutral polarity  have been tagged with a value equal to $0$.

Two human annotators have been involved in the tagging process. The whole annotation process took more than three months.

\begin{figure}[t!]
\begin{center}
\includegraphics[width=0.99\textwidth]{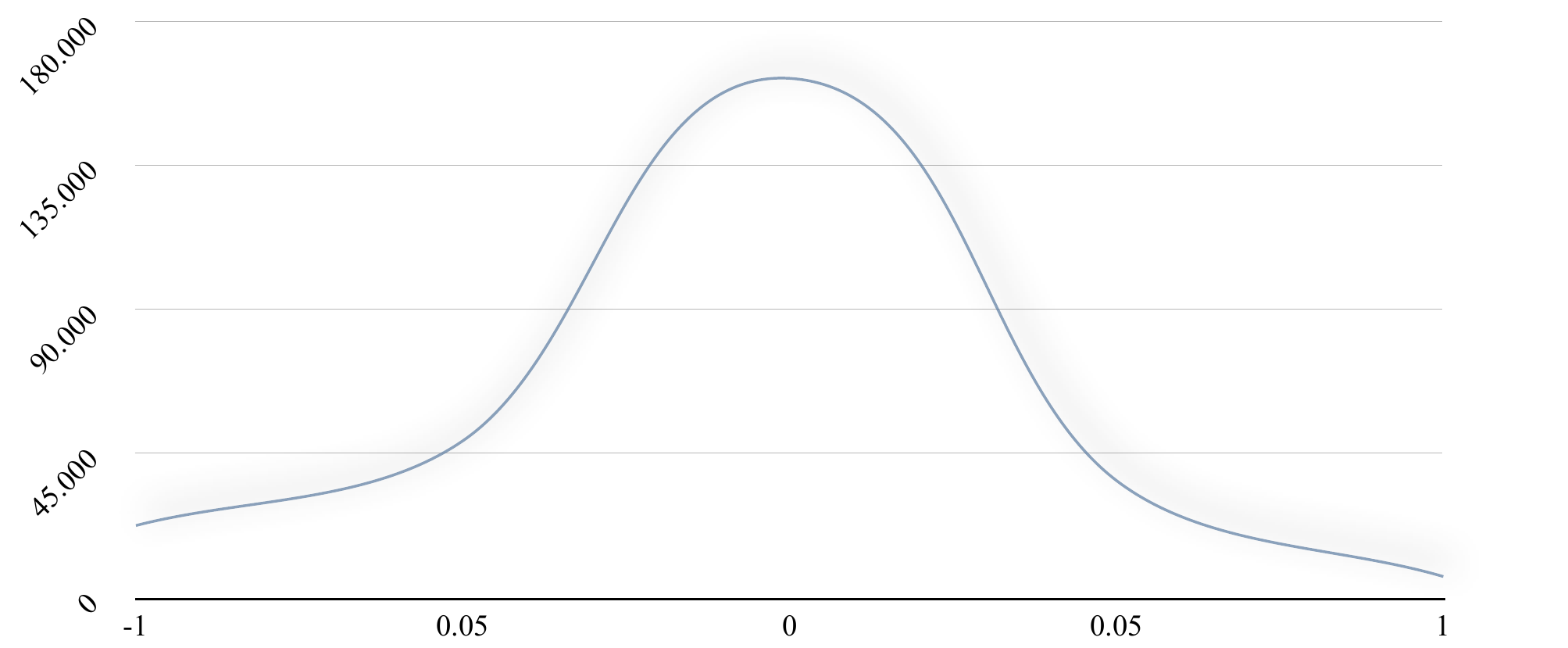}
\end{center}
\caption{\label{fig:distribution}The polarity distribution of the 277.387 different words of the Ispell Italian dictionary. Words are tagged with five different polarity values between $-1.0$ and $+1.0$.}
\end{figure}

Figure \ref{fig:distribution} shows the polarity distribution of all words of the Italian dictionary. We observed $162.000$ words which have been tagged with a neutral sentiment polarity, more than $70.000$ with a negative polarity and more than $43.000$ words tagged with a positive polarity.

Specifically words evoking a negative sentiment are divided in two sets, $22.651$ with a strongly negative polarity and $49.074$ words with a fairly negative polarity.
Similarly, in the case of words evoking a positive sentiment, we observed $6.739$ words with a strongly positive polarity and $36.688$ words with a fairly positive polarity.
Table \ref{tab:distribution} shows in details the number of words detected for each polarity value together with the percentage of words detected in each group. Notice that more than $40\%$ of words have been assigned to a polarity values, while $58\%$ of words have been assigned with a neutral polarity.

\begin{table}
\begin{center}
	\begin{tabular*}{0.85\textwidth}{@{\extracolsep{\fill} }llcc}
	\hline
	&&&\\[-0.2cm]
	\textbf{polarity}  &\textbf{value} & \textbf{\# of words} & \textbf{\% of words}\\[0.2cm]
	\hline
	&&&\\[-0.2cm]
	strongly negative  &$-1.0$	 & $22.651$ & $8,17\%$\\
	negative &$-0.5$ & $49.074$ & $17,70\%$\\
	neutral & $\pm0.0$ & $162.170$ & $58,47\%$\\
	positive &$+0.5$ & $36.688$  & $13,23\%$\\
	strongly positive &$+1.0$ & $6.739$ & $2,43\%$\\[0.2cm]
	\hline
	\end{tabular*}
\end{center}
\caption{\label{tab:distribution}The distribution of polarity values assigned to Italian words.}
\end{table}

\begin{table}
\begin{center}
	\begin{tabular*}{0.47\textwidth}{@{\extracolsep{\fill} }lll}
	\multicolumn{3}{l}{\textsc{Strongly Negative Polarity}}\\[0.2cm]
	\hline
	&&\\[-0.2cm]
	\textbf{word}  & \textbf{type} & \textbf{translation} \\[0.2cm]
	\hline
	&&\\[-0.2cm]
	\expl{castigo} & noun &  \emph{penance, scourge}\\
	\expl{cattivo} & adj. & \emph{bad}\\
	\expl{cedevole} & adj. & \emph{weak, soft}\\
	\expl{celare} & verb & \emph{conceal, hide} \\
	\expl{cialtrone} & adj. & \emph{boor, lout}\\
	\expl{ciarlatano} & adj. & \emph{barker, charlatan}\\
	\expl{cinico} & adj. & \emph{cynical} \\
	\expl{ciuco} & noun & \emph{jackass} \\
	\expl{coatto} & adj. & \emph{forced} \\
	\expl{cocci} & noun & \emph{pieces, potsherd} \\[0.2cm]
	\hline
	\end{tabular*}~~~~~~
	\begin{tabular*}{0.47\textwidth}{@{\extracolsep{\fill} }lll}
	\multicolumn{3}{l}{\textsc{Negative Polarity}}\\[0.2cm]
	\hline
	&&\\[-0.2cm]
	\textbf{word}  & \textbf{type} & \textbf{translation} \\[0.2cm]
	\hline
	&&\\[-0.2cm]
	\expl{catapecchia} & noun &  \emph{hovel, shanty}\\
	\expl{catorcio} & noun & \emph{untrustworthy car}\\
	\expl{catturare} & verb & \emph{capture, catch}\\
	\expl{causare} & verb & \emph{cause, provoke} \\
	\expl{caustico} & adj. & \emph{corrosive, caustic}\\
	\expl{cavia} & noun & \emph{test subject}\\
	\expl{cella} & noun & \emph{prison cell} \\
	\expl{cencio} & noun & \emph{rag, wreck} \\
	\expl{cessare} & verb & \emph{cease, stop} \\
	\expl{chiacchierone} & adj. & \emph{chatty, prattler} \\[0.2cm]
	\hline
	\end{tabular*}
\end{center}
\caption{\label{tab:negative}Examples of words from the Italian dictionary which have been tagged with a strongly negative polarity (on the left) and with a negative polarity (on the right).}
\end{table}

\begin{table}
\begin{center}
	\begin{tabular*}{0.47\textwidth}{@{\extracolsep{\fill} }lll}
	\multicolumn{3}{l}{\textsc{Strongly Positive Polarity}}\\[0.2cm]
	\hline
	&&\\[-0.2cm]
	\textbf{word}  & \textbf{type} & \textbf{translation} \\[0.2cm]
	\hline
	&&\\[-0.2cm]
	\expl{campione} & noun &  \emph{champion}\\
	\expl{capace} & adj. & \emph{able, competent}\\
	\expl{carina} & adj. & \emph{cute, pretty}\\
	\expl{carezza} & noun & \emph{caress, touch} \\
	\expl{celebre} & adj. & \emph{famous, well-known}\\
	\expl{chiarezza} & noun & \emph{clarity, frankness}\\
	\expl{chic} & adj. & \emph{chic, classy} \\
	\expl{coccole} & noun & \emph{snuggles, ciddles} \\
	\expl{collaborare} & verb & \emph{collaborate} \\
	\expl{colto} & adj. & \emph{cultured} \\[0.2cm]
	\hline
	\end{tabular*}~~~~~~
	\begin{tabular*}{0.47\textwidth}{@{\extracolsep{\fill} }lll}
	\multicolumn{3}{l}{\textsc{Positive Polarity}}\\[0.2cm]
	\hline
	&&\\[-0.2cm]
	\textbf{word}  & \textbf{type} & \textbf{translation} \\[0.2cm]
	\hline
	&&\\[-0.2cm]
	\expl{castello} & noun &  \emph{castle, fortress}\\
	\expl{cautela} & noun & \emph{caution, prudence}\\
	\expl{cavaliere} & noun & \emph{knight, gentleman}\\
	\expl{cavarsela} & verb. & \emph{cause, provoke} \\
	\expl{celere} & adj. & \emph{manage, hack it}\\
	\expl{celeste} & adj. & \emph{sky-blue, celestial}\\
	\expl{centrare} & verb. & \emph{center, hit the target} \\
	\expl{champagne} & nm. & \emph{champagne} \\
	\expl{chicca} & noun & \emph{tidbit, treat} \\
	\expl{cinguettio} & noun & \emph{twitter, chirping} \\[0.2cm]
	\hline
	\end{tabular*}
\end{center}
\caption{\label{tab:positive}Examples of words from the Italian dictionary which have been tagged with a strongly positive polarity (on the left) and with a positive polarity (on the right).}
\end{table}

\subsection{Polarity Modifiers} \label{sec:mod}

An adjective is a word or set of words that modifies a noun or a pronoun. In most cases adjectives come before the word they modify.
Some adjective can modify the polarity of a noun with a non neutral prior polarity. For example the adjective \emph{raro} (\emph{rare}) can be used in composition with the adjective \emph{bellezza} (\emph{beauty}) to emphasize its positive meaning (\emph{a women with a rare beauty}).
Similarly the adjective \emph{esiguo} (\emph{scarse}) can be used in combination with the noun \emph{valore} (\emph{virtue}) changing its positive polarity in a negative sentiment (\emph{a man with scarse virtue}).

An adverb is a word or set of words that modifies verbs, adjectives, or other adverbs.
Generally an adverb answers how, when, where, or to what extent an action is performed or an adjective is applicable.
In this context some adverbs are able to modify the sentiment evoked by a verb or by an adjective with non neutral polarity.
For instance the adverb \emph{appena} (\emph{barely}) can be associated with an adjective in order to reduce its positive (or negative) polarity, e.g. \emph{barely succeed} or \emph{barely enthusiast}.
Similarly the adverb \emph{davvero} (\emph{truly}) can be associated with an adjective like \emph{sorprendente} (\emph{amazing}) in order to emphasize its positive meaning.

In our work we collected a set of more than $200$ polarity modifier which have been manually tagged with a proportionality factor ranging between $-2.0$ and $+2.0$. When a term with a non neutral polarity $x$ is associated with a modifier with a proportionality factor $y$, we obtain a compound term whose polarity can be estimated as $(x\times y)$. 

Depending on the value of such factor we can distinguish four different kind of modifiers.
\begin{itemize}
\item \textbf{Emphasize}\\
These modifiers have a proportionality factor greater than $+1.0$ and, when associated with a term having a non neutral polarity, evokes a sentiment which is stronger than the original one. thus they emphasize a positive (or negative) polarity value.\\
\begin{quote}
	\expl{proprio bello} (\emph{really beautiful}) $= +1.6 \times  +1.0 = +1.6$\\
	\expl{alquanto sgradevole} (\emph{rather unpleasant}) $= +1.5 \times  -1.0 = -1.5$\\
	\expl{grande valore} (\emph{great virtue}) $= +1.8 \times +0.5 = +0.9$\\
\end{quote}
\item \textbf{Moderate}\\
These modifiers have a proportionality factor greater than $0.0$ and smaller than $+1.0$. When associated with a term having a non neutral polarity, they result in a compound term with a moderated sentiment which is weaker than the original one.\\
\begin{quote}
	\expl{appena vinto} (\emph{just gained}) $= +0.7 \times  +0.5 = +0.35$\\
	\expl{mediamente brutto} (\emph{ugly on average}) $= +0.5 \times  -1.0 = -0.5$\\
	\expl{breve successo} (\emph{brief success}) $= +0.6 \times +0.5 = +0.3$\\
\end{quote}
\item \textbf{Reverse and moderate}\\
This kind of modifiers have a proportionality factor greater than $-1.0$ and smaller than $0.0$. When they are associated with a term having a non neutral polarity, evoke a sentiment which is in opposition with the original sentiment, but has an absolute value of polarity which is smaller than the original polarity.\\
\begin{quote}
	\expl{poco ragionevole} (\emph{little reasonable}) $= -0.7 \times  +0.5 = -0.35$\\
	\expl{esiguo dolore} (\emph{scarse pain}) $= -0.7 \times  -1.0 = +0.7$\\
	\expl{limitato guadagno} (\emph{limited benefit}) $= -0.8 \times +1.0 = -0.8$\\
\end{quote}
\item \textbf{Reverse and emphasize}\\
These modifiers have a proportionality factor smaller or equal than $-1.0$ and, if associated with a term having a non neutral polarity, evokes a sentiment which is stronger than the original one but with an opposite polarity.\\
\begin{quote}
	\expl{insufficiente prestigio} (\emph{insufficient prestige}) $= -1.2 \times  1.0 = -1.2$\\
	\expl{minime scomodità} (\emph{minimal inconvenience}) $= -1.0 \times  -0.5 = +0.5$\\
	\expl{scarso valore} (\emph{lacking virtue}) $= -1.2 \times +0.5 = -0.6$\\
\end{quote}
\end{itemize}

\begin{table}
\begin{center}
	\begin{tabular*}{0.45\textwidth}{@{\extracolsep{\fill} }lll}
	\multicolumn{3}{l}{\textsc{Positive Polarity Modifiers}}\\[0.2cm]
	\hline
	&&\\[-0.2cm]
	\textbf{word}  & & \textbf{translation} \\[0.2cm]
	\hline
	&&\\[-0.2cm]
	alquanto & $+1.5$ &  \emph{rather}\\
	appena & $+0.7$ & \emph{barely}\\
	parecchio & $+2.0$ & \emph{much, a lot}\\
	abbastanza & $+1.7$ & \emph{sufficiently} \\
	tanto & $+2.0$. & \emph{very}\\
	ovviamente & $+1.8$. & \emph{obviously}\\
	moderatamente & $+0.5$ & \emph{fairly} \\
	mediamente & $+0.7$ & \emph{on average} \\
	proprio & $+1.6$ & \emph{really} \\
	davvero & $+2.0$ & \emph{truly} \\[0.2cm]
	\hline
	\end{tabular*}~~~~~~
	\begin{tabular*}{0.45\textwidth}{@{\extracolsep{\fill} }lll}
	\multicolumn{3}{l}{\textsc{Negative Polarity Modifiers}}\\[0.2cm]
	\hline
	&&\\[-0.2cm]
	\textbf{word}  & & \textbf{translation} \\[0.2cm]
	\hline
	&&\\[-0.2cm]
	affatto 		& $-1.5$ &  \emph{not at all}\\
	neppure 		& $-1.0$ & \emph{not even}\\
	poco 		& $-0.7$ & \emph{little}\\
	scarsamente 	& $-0.8$ & \emph{poorly} \\
	esiguo 		& $-0.7$ & \emph{scarse, scant}\\
	minimo 		& $-1.0$ & \emph{smallest}\\
	limitato 		& $-0.8$ & \emph{limited} \\
	insufficiente 	& $-1.2$ & \emph{insufficient} \\
	scarso 		& $-1.2$ & \emph{lacking} \\
	corto 		& $-0.6$ & \emph{short} \\[0.2cm]
	\hline
	\end{tabular*}
\end{center}
\caption{\label{tab:modifiers}Some examples of polarity modifiers and their respective proportionality factors: (on the left) positive polarity modifiers and (on the right) negative polarity modifiers.}
\end{table}

In Table \ref{tab:modifiers} some examples of polarity modifiers are shown together with their respective proportionality factors.

\section{A Web Based Frontend} \label{sec:frontend}
We implemented a simple web based tool in order to access the lexical resource presented in this paper.
In order to allow a blind review of the paper we uploaded the tool in a free hosting server. The tool is accessible at the url
\begin{quote} 
\url{http://www.dmi.unict.it/~faro/sabrina/} 
\end{quote}
The tool allows to evaluate single Italian terms or compound terms, where words with a non neutral polarity are associated with modifiers, as described above.
Moreover each example which you can find above in the paper is tagged with an anchor which redirect the reader to the web page of the tool in order to evaluate the sentiment value of the example itself.

If a whole sentence is tested by the tool, containing more than one term with non neutral prior polarity, then a straightforward approach is applied in order to compute an approximation of the polarity of the whole sentence.
In particular the set of polarity values contained in the sentence is arranged from the lowest one to the highest one and the median of such a set is taken as the polarity value of the whole sentence.

Specifically the median is the number separating the higher half of the set of polarity values from the lower half. If there is an even number of polarity values, then there is no single middle value. Int this cases the median is usually defined to be the mean of the two middle values.

\section{Conclusions} \label{sec:conclusions}
In this paper we presented a new lexical resource for the Italian language containing more than 277.000 words which have been manually tagged with their prior polarity values, i.e. a value indicating the sentiment which such words evoke when are out of any context. We also provide an additional lexical resource containing a set of more than 200 polarity modifiers which can be used for inducing the sentiment polarity of Italian compound terms.
Future works will be devoted to test the effectiveness of such resource in opinion mining task.

\bibliographystyle{plain}

\end{document}